\def\year{2022}\relax
\documentclass[letterpaper]{article} 
\usepackage{aaai22}  
\usepackage{times}  
\usepackage{helvet}  
\usepackage{courier}  
\usepackage[hyphens]{url}  
\usepackage{graphicx} 
\usepackage{natbib}  
\usepackage{caption} 
\DeclareCaptionStyle{ruled}{labelfont=normalfont,labelsep=colon,strut=off} 
\usepackage{algorithm}
\usepackage{algorithmic}
\usepackage{hyperref}
\usepackage{glossaries}
\usepackage{amsmath}
\usepackage{amssymb}
\usepackage{mathtools}
\usepackage{amsthm}

\usepackage{microtype}
\usepackage{graphicx}

\usepackage[capitalize,noabbrev]{cleveref}

\usepackage{newfloat}
\usepackage{listings}
\floatstyle{ruled}
\newfloat{listing}{tb}{lst}{}
\floatname{listing}{Listing}
\title{Tunable Complexity Benchmarks for Evaluating Physics-Informed Neural Networks on Coupled Ordinary Differential Equations}
\author{
    Alexander New\textsuperscript{\rm 1}, Benjamin Eng\textsuperscript{\rm 1}, Andrea C. Timm\textsuperscript{\rm 1}, Andrew S. Gearhart\textsuperscript{\rm 1}
}
\affiliations{
    \textsuperscript{\rm 1}Research and Exploratory Development Department, Johns Hopkins University Applied Physics Laboratory\\

    11100 Johns Hopkins Road Laurel\\
    Maryland 20723\\
    alex.new@jhuapl.edu
}

\glsdisablehyper
\newacronym{APL}{APL}{%
  the Johns Hopkins University Applied Physics Laboratory%
}

\newacronym{BC}{BC}{boundary condition}

\newacronym{CFD}{CFD}{computational fluid dynamics}

\newacronym{DNS}{DNS}{direct numerical simulation}
\newacronym{DE}{DE}{differential equation}

\newacronym{FFD}{FFD}{fast fluid dynamics}

\newacronym{FEM}{FEM}{finite element method}

\newacronym[plural=GPUs,longplural={graphics processing units}]{GPU}{GPU}{%
  graphics processing unit%
}

\newacronym{IC}{IC}{initial condition}

\newacronym{JHU}{JHU}{%
  Johns Hopkins University%
}

\newacronym{ML}{ML}{machine learning}
\newacronym{MLP}{MLP}{multi-layer perceptron}

\newacronym{NS}{NS}{Navier-Stokes}
\newacronym{NSE}{NSE}{Navier-Stokes equations}
\newacronym{NN}{NN}{neural network}
\newacronym{NTK}{NTK}{neural tangent kernel}

\newacronym{OR}{OR}{operating room}
\newacronym{ODE}{ODE}{ordinary differential equation}

\newacronym[
    shortplural={PINNs}
]{PINN}{PINN}{physics-informed neural network}
\newacronym{PDE}{PDE}{partial differential equation}
\newacronym{RANS}{RANS}{Reynolds-averaged Navier-Stokes}

\newacronym{SHM}{SHM}{simple harmonic motion}

\newacronym{SF}{SF}{stream-function}

\newacronym{VP}{VP}{velocity-pressure}
\newacronym{VV}{VV}{velocity-velocity}

\begin{document}

\maketitle

\begin{abstract}
In this work, we assess the ability of \glspl{PINN} to solve increasingly-complex coupled \glspl{ODE}. We focus on a pair of benchmarks: discretized partial differential equations and harmonic oscillators, each of which has a tunable parameter that controls its complexity. Even by varying network architecture and applying a state-of-the-art training method that accounts for ``difficult'' training regions, we show that \glspl{PINN} eventually fail to produce correct solutions to these benchmarks as their complexity---the number of equations and the size of time domain---increases. We identify several reasons why this may be the case, including insufficient network capacity, poor conditioning of the \glspl{ODE}, and high local curvature, as measured by the Laplacian of the \gls{PINN} loss.
\end{abstract}

\glsresetall

\section{Introduction}
In recent years, there has been much interest in using \gls{ML} models to approximate systems governed by mechanistic dynamics~\cite{willard2020integrating,karniadakis2021physics}. 
This includes rapid forward simulations and the ability to determine unknown components of dynamics via solution of inverse problems. Typically, these physics-based \gls{ML} models incorporate \textit{a priori} knowledge of the system to be modeled---via a black-box simulation environment or an explicit system of \glspl{DE}. In this work, we analyze a popular approach to the latter strategy: \glspl{PINN}~\cite{raissi2019physics}. \Glspl{PINN} incorporate \glspl{DE} as physics-based regularizers in the objective function of a neural network (i.e., soft constraints) along with terms associated with initial and boundary conditions. This is a straightforward approach to incorporating known dynamics into neural networks and showed early success~\cite{mclenny2020adaptive,lu2021deepxde}, but recent results~\cite{mclenny2020adaptive,krishnapriyan2021characterizing,wang2021understanding,wang2022ntk} highlight difficulties that \glspl{PINN} face due to their complex objectives. Our analysis builds upon these latter efforts, focusing on representing coupled systems of \glspl{ODE}.

\Gls{ODE} systems, or differential equations with a single independent variable and one or more dependent variables, arise in a number of use cases. These include modeling reaction-diffusion processes in synthetic biology~\cite{singhal2021txtlsim} and chemical kinetics~\cite{Ji2021pinns_chemistry}, as well as solving \glspl{PDE} via the method of lines~\cite{Verwer1984methodoflines}. This leads, for example, to applications in fluid mechanics like Burgers' equation~\cite{Zogheib2021burgers} and condensed matter physics such as lattice dynamics~\cite{chow1996dynamics}.

Systems of~\glspl{ODE} vary in complexity based on domain size, number of equations being modeled, and problem stiffness. These challenges occur regularly---for example, when discretizing \glspl{PDE} into coupled \glspl{ODE}, the number of \glspl{ODE} scales with discretization granularity. This is a computational bottleneck in higher-fidelity simulations. Due to the ubiquity of these systems, it is natural to explore how effectively \glspl{PINN} solve them, especially as their complexities increase. Similar work was performed by~\cite{krishnapriyan2021characterizing}, who use underlying \gls{PDE} parameters such as convection coefficients as proxies for the complexity of the learning problem, and show that \glspl{PINN} fail to solve these problems as the complexity increases.


In this paper, we review the formulation of \glspl{PINN} (\cref{sec:pinns}) and introduce two problems that involve systems of coupled \glspl{ODE} (\cref{sec:studies}). Each problem is characterized by a tunable ``complexity''. We then demonstrate that \glspl{PINN}, even with the use of a state-of-the-art training method, fail to produce correct solutions to the system of \glspl{ODE} (as indicated by a classical \gls{ODE} solver) as the complexity of the problem increases (\cref{sec:results}). We then link the intuitive complexity of the problem with a quantitative characterization of the difficulty via the Laplacian of the \gls{PINN} learning problem.


\section{Approach}\label{sectapproach}

\subsection{Physics-Informed Neural Networks}\label{sec:pinns}

\glspl{PINN}~\cite{raissi2019physics} are neural networks with weights trained such that the network satisfies a differential equation.  The use of a neural network to satisfy a \gls{DE} is motivated by the universal approximation theorem;
however, this theorem does not guarantee that a specific optimization procedure will yield a network satisfying the \gls{DE}. \cite{shin2020convergence} showed that, for certain categories of \gls{PDE}, as the amount of available data increases, the \gls{PINN} learning problem will converge to a solution to the \gls{PDE}.

Due to the overhead of training, solving a \gls{DE} with a \gls{PINN} takes longer than using a classical numerical method. Unlike classical methods, however, \glspl{PINN} are easily extended to inverse problems---i.e., given a partially-specified governing equation and some data, recover the full governing equation. Furthermore, \glspl{PINN} do not require a mesh, which enables efficient application to new problems. The forward prediction setting, which we consider here, is a different use-case than many \gls{ML} applications, in that the training signal uses a \gls{DE} (or system of \glspl{DE}), rather than a labelled dataset. 

To define \glspl{PINN}, consider a general \gls{ODE} that determines function $\mathbf{u} : \mathbb{R} \to \mathbb{R}^N$:
\begin{eqnarray*}
    \mathcal{N}(\mathbf{u})(t) &=& 0,\,\,\,\,\,t\in(0, T)\\
    \mathcal{I}(\mathbf{u})(t) &=& 0,\,\,\,\,\,t = 0,
\end{eqnarray*}
where $\mathcal{N}$ is a differential operator, and $\mathcal{I}$ captures deviation from prespecified initial conditions. For the first-order \glspl{DE} considered here, $\mathcal{I}(\mathbf{u})(t) = \mathbf{u}(t) - \mathbf{u}_0$, where $\mathbf{u}_0$ is fixed. 
Then a \gls{PINN} $\hat{\mathbf{u}}_w: \mathbb{R} \to \mathbb{R}^N$ is a neural network parameterized by 
a weight vector $w \in \mathbb{R}^M$ that satisfies
\begin{equation}
    \min_w L(w) = \sum_{t_d \in \mathcal{T}} ||\mathcal{N}(\hat{\mathbf{u}}_w)(t_d)||^2 + \nu_\mathcal{I}||\mathcal{I}(\mathbf{\hat{u}}_w)(0)||^2,
    \label{eq:base_pinn}
\end{equation}
where $\nu_{\mathcal{I}} > 0$ weights the initial condition components and $\mathcal{T} = \{t_1, \hdots, t_D\} \subseteq (0, T)$ is the set of training points. The \gls{PINN} literature typically refers to the loss term involving the differential operator $\mathcal{N}$ as the residual loss.

The \gls{PINN} loss $L(w)$ may be minimized with first-order methods like Adam~\cite{Kingma2014adam}.
Unlike many other \gls{ML} settings, it includes derivatives of $\hat{\mathbf{u}}_w$ with respect to its input $t$. These may be calculated using the same automatic differentiation procedures~\cite{Baydin2017autodiff} used to calculate loss gradients with respect to network weights, but this means that the weight update of a \gls{PINN} loss function contains second- and higher-order derivatives (a gradient calculation with respect to $w$, combined with derivatives with respect to $t$ based on the \gls{DE}).

In practice, we often solve a modified formulation of \cref{eq:weighted_pinn}, introduced by~\cite{mclenny2020adaptive}:
\begin{equation}
    \min_w \max_\lambda L(w, \lambda)
    \label{eq:weighted_pinn}
\end{equation}
where 
\begin{eqnarray*}
L(w, \lambda) &=& \frac{1}{|\mathcal{T}|}\sum_{t_d \in \mathcal{T}} \mu(\lambda_d) ||\mathcal{N}(\mathbf{\hat{u}}_w)(t_d)||^2 \\&& + \mu(\lambda_0) \nu_{\mathcal{I}}||\mathcal{I}(\mathbf{\hat{u}}_w)(0)||^2,
\end{eqnarray*}
where $\lambda \in \mathbb{R}^{D+1}_{\geq 0}$ are attention weights associated with each point $t_d \in \{0\}\cup\mathcal{T}$, and $\mu: \mathbb{R}^+ \to \mathbb{R}^+$ is a masking function required to be differentiable, non-negative, and strictly increasing. This problem can be solved with a joint gradient ascent (for attention weights $\lambda$) and descent (for \gls{PINN} weights $w$) procedure.



\cite{krishnapriyan2021characterizing} observe that the \gls{PINN} learning problem can be ill-conditioned due to the differential operator $\mathcal{N}$ in $L$. We continue this line of analysis by considering the Laplacian $\Delta L_c$ of a component $c$ (residual loss or initial condition loss) of $L$ with respect to the network weights $w$:
$$\Delta L_c(w) = \sum_{p} \partial_{w_p,w_p} L_c(w),$$
where $L_c$ is the residual loss or the initial condition loss in~\cref{eq:base_pinn}.
The Laplacian's magnitude measures the local size of the learning problem's curvature. We evaluate the Laplacian by estimating the trace of the Hessian $\nabla^2 L_c(w)$ using Hutchinson's method~\cite{yao2020pyhessian}:
$$\Delta L_c(w) = \mathrm{tr}\,\nabla^2 L_c(w) = \mathbb{E}_v\{v^T [(\nabla^2 L_c)(w)]v\},$$
where $v \in \mathbb{R}^M$ has $iid$ components sampled from a Rademacher distribution, and the Hessian-vector products do not require the full Hessian~\cite{pearlmutter1994hessian}.


\glspl{PINN} have primarily been used to solve and analyze \glspl{PDE} rather than \glspl{ODE}. This is partially because, unlike for \glspl{PDE}~\cite{brandstetter2022message}, classical \gls{ODE} solvers such as Runge-Kutta~\cite{Dormand1980rk45} are efficient for forward predictions of \gls{ODE} systems. We focus here on forward prediction of \glspl{ODE} because they represent the simplest-possible application domain of \glspl{PINN}. \Gls{PINN} failures during forward prediction should be addressed to allow general use of \glspl{PINN} in other settings like inverse problems and \glspl{PDE}.

\subsection{The \texttt{pinn-jax} library}\label{sec:library}

\gls{PINN} research has benefited from several open-source libraries, including DeepXDE~\cite{lu2021deepxde}, NeuralPDE.jl~\cite{zubov2021neuralpde}, and TensorDiffEq~\cite{mcclenney2021tensordiffeq}. DeepXDE in particular easily enables solving \glspl{PINN} over irregular domain geometries. For this work, we implemented our \glspl{PINN} in a new library, \texttt{pinn-jax}, which is built on the \texttt{jax}\footnote{\url{https://github.com/google/jax}} framework and uses \texttt{flax}\footnote{\url{https://github.com/google/flax}} for neural network layers and \texttt{optax}\footnote{\url{https://github.com/deepmind/optax}} for optimization.
The use of \texttt{jax} improves code performance and allows calculations to be run on CPUs/GPUs/TPUs as available, while preserving the flexibility to specify different \glspl{ODE} and obtain their solutions via different \gls{PINN} training strategies (e.g., different weighting schemes) and neural network architectures (e.g., \glspl{MLP} and ResNets~\cite{He2016residual}).

As we consider scalar inputs for our networks, \texttt{pinn-jax} implements first-order loss derivatives with forward-mode automatic differentiation (the \texttt{jacfwd} function), and second-order loss derivatives with forward-over-forward-mode automatic differentiation. The \texttt{vmap} function enables efficient calculation of these derivatives over an entire batch of domain points. Calculation of Hessian-vector products to assess loss Laplacians $\Delta L_c$ is accomplished by composing the Jacobian-vector product function \texttt{jvp} with the gradient function \texttt{grad}. Although we focus on \glspl{PINN} for \glspl{ODE}, we note that \texttt{pinn-jax} also includes functionality for solving \glspl{PDE}. In that component of the library, we use DeepXDE's ~\cite{lu2021deepxde} \texttt{geometry} module, which enables specification of and sampling from complex domains.

\subsection{Benchmarks}\label{sec:studies}
After finding it difficult to train \glspl{PINN} on a set of biological reaction equations, we decided to explore the broader problem of using \glspl{PINN} to solve systems of \glspl{ODE}. This resulted in the development of two benchmarks with parameterized complexity---one that scales the number of equations in the system (a discretized heat equation) and the other that defines complexity as a maximum simulation time (\gls{SHM}). Each of our benchmarks has additional parameters (e.g., choice of initial conditions) that we do not vary in these results---our goal is to leave much fixed and explore the effect of increasing complexity on the ability of a \gls{PINN} to solve these \gls{ODE} systems.

\subsubsection{Discretized heat equation}\label{sec:heat}

The heat equation, $\mathbf{u}_t = \Delta \textbf{u}$, is defined for a scalar field $u$ on the domain $(x, t) \in (0, 1)\times(0, T)$. In particular, when sources are added to it, one obtains a broad family of reaction-diffusion systems applicable to lattice problems~\cite{chow1996dynamics}. Although the heat equation may be solved as a \gls{PDE} with a \gls{PINN}, here we explore how effectively \glspl{PINN} solve it after it is discretized into a set of coupled \glspl{ODE}. In particular, we apply the method of lines~\cite{Verwer1984methodoflines} and discretize the spatial dimension into $N$ evenly-spaced points $x_1,\hdots,x_N$, where $x_n = (n - 1) \Delta x= (n - 1) / (N - 1)$, and then approximate the second derivative with a central finite difference. This yields the following coupled \gls{ODE}, for a function $\mathbf{u} : \mathbb{R} \to \mathbb{R}^N$:
\begin{eqnarray*}
    \dot{\mathbf{u}} &=& \mathbf{A}\mathbf{u} + \mathbf{f},\,\,\,\,t\in(0, T),\\
    \mathbf{u}_{t=0} &=& [g(x_1), \hdots, g(x_N)]^T
\end{eqnarray*}
where the matrix $\mathbf{A}$ is given by    
\begin{eqnarray*}
    \mathbf{A}_{n,n} &=& -2(N-1)^2\\
    \mathbf{A}_{n,n+1} &=& (N-1)^2\\
    \mathbf{A}_{n+1,n} &=& (N-1)^2\\
    \mathbf{A}_{n,n'} &=& 0.
\end{eqnarray*}
Further, $\mathbf{f} = (N-1)^2[u_L(t), 0, \hdots, 0, u_R(t)]^T$, $g$ specifies the initial condition, and $u_L$ and $u_R$ specify the boundary conditions on the left and right side, respectively, of the spatial domain $(0, 1)$.

\begin{figure}
    \centering
    \includegraphics[width=\linewidth]{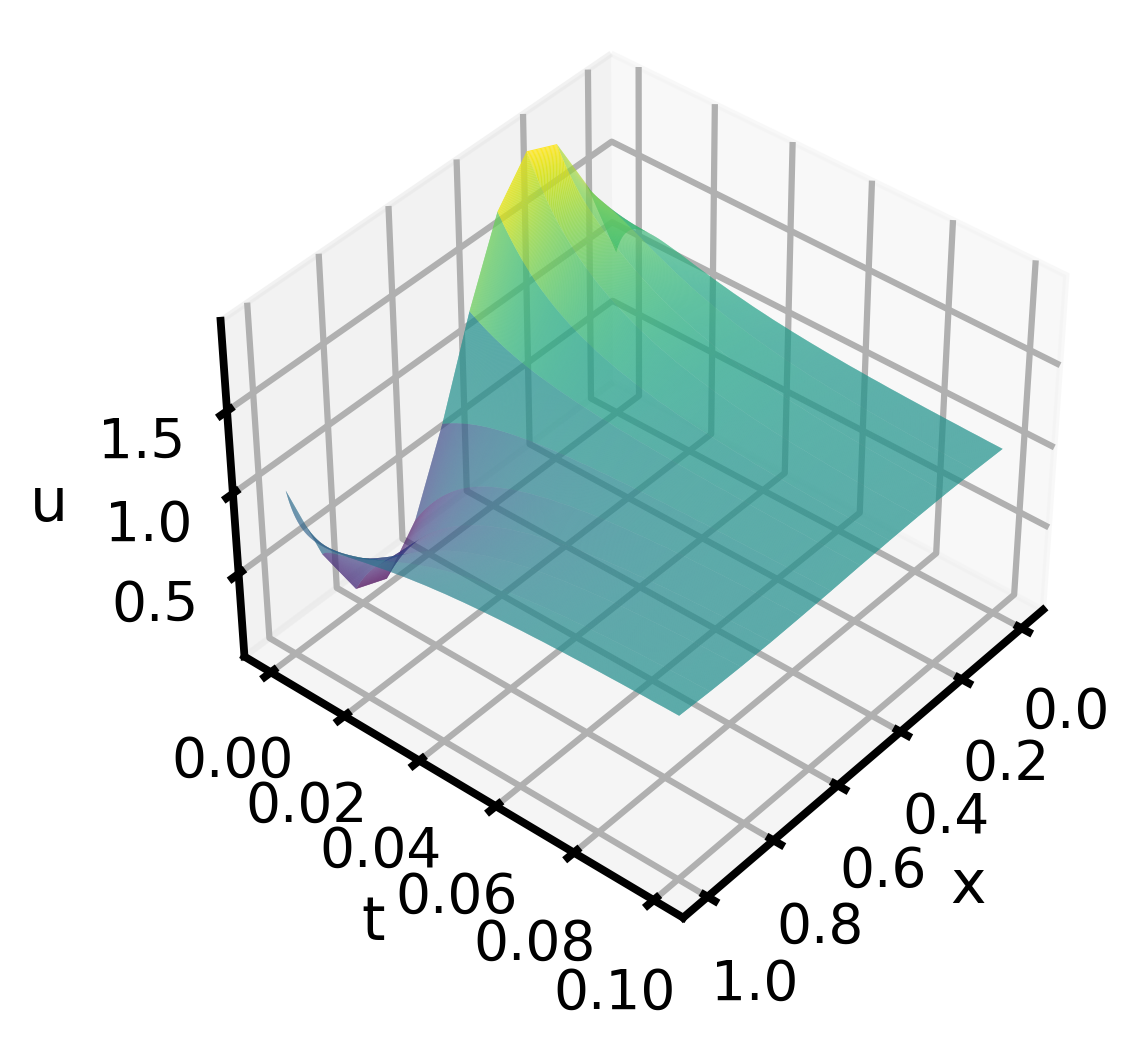}
    \caption{An example solution to the heat equation (\cref{sec:heat}). At $t=0$, the system has a sinusoidal shape; as $t$ increases, the solution decays to a constant value. As the discretization of the spatial dimension increases, the problem becomes more ill-conditioned.}
    \label{fig:example_heat}
\end{figure}

An example solution to this problem is shown in~\cref{fig:example_heat}. In this discretized formulation, the complexity of the problem is determined by the size of the discretization $N$.\footnote{We choose to keep the maximum time $T$ constant for the heat equation benchmark to focus on equation scaling.} The discretized differential operator $\mathbf{A}$ couples each state variable $u_n$ to its spatially-adjacent variables $u_{n+1}$ and $u_{n-1}$.

As the number of discretized space-points $N$ increases, the coupled \glspl{ODE} become more ill-conditioned. Specifically, as $\mathbf{A}$ is a symmetric tridiagonal Toeplitz matrix, it has a closed-form representation of its eigenvalues~\cite{Noschese2013toeplitz}:
\begin{eqnarray*}
    e_n &=& -2(N-1)^2(1 - \cos(n\pi/(N+1))),
\end{eqnarray*}
for $n=1,\hdots,N$.

In this case, the condition number is given by:
$$\kappa_N = \frac{|e_N|}{|e_1|} = \frac{1 - \cos (N\pi / (N+1))}{1 - \cos (\pi / (N+1))}$$
Since the condition number increases without bound as the discretization size $N$ increases, we have an explicit connection between an intuitive notion of complexity ($N$) and a more formal notion of complexity (the condition number of the matrix $\mathbf{A}$). Further details on this benchmark are found in~\cref{sec:heat_deets}.

\subsubsection{Long time-horizon simple harmonic motion}\label{sec:sinusoids}
\begin{figure}
    \centering
    \includegraphics[width=0.9\linewidth]{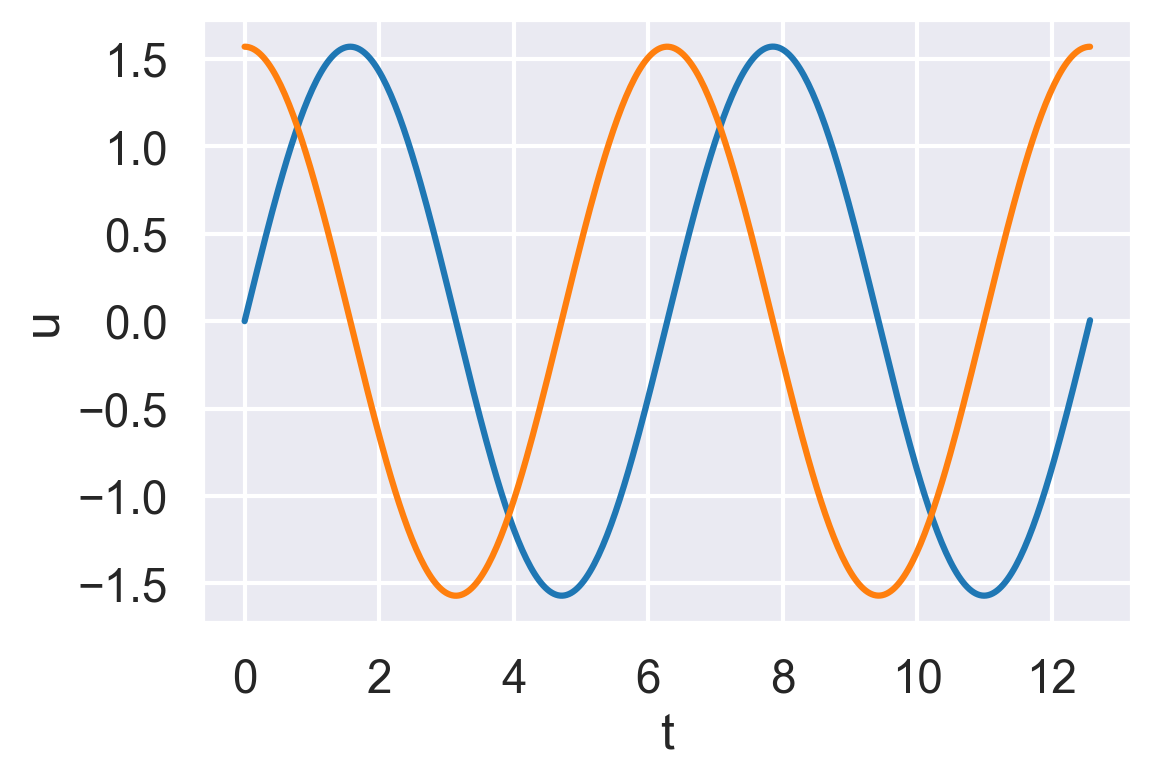}
    \caption{A solution to the \gls{SHM} problem (\cref{sec:sinusoids}); as the size of the time domain increases, the model becomes harder to solve with a \gls{PINN}. The initial condition $\mathbf{u}_0 = [0, \pi/2]^T$ was used with a maximum time of $T = 4\pi$ and a frequency of $\omega = 1$.}
    \label{fig:example_sinusoid}
\end{figure}

\glsreset{SHM}
We begin by considering a classic coupled \gls{ODE}: \gls{SHM}, and we increase its complexity by evaluating it over increasingly large time domains. This \gls{ODE}, for a function $\mathbf{u} : \mathbb{R} \to \mathbb{R}^2$, is specified by
\begin{eqnarray*}
    \dot{\mathbf{u}} &=& \mathbf{A} \mathbf{u},\,\,\,\,\,\,\,\,\,\,t \in (0, T)\\
    \mathbf{u}_{t=0} &=& \mathbf{u}_0
\end{eqnarray*}
where the matrix $\mathbf{A}$ is given by $\mathbf{A}_{1,2} = -\omega$, $\mathbf{A}_{2,1} = \omega$, and $\mathbf{A}_{1,1} = \mathbf{A}_{2,2} = 0$, for a frequency $\omega > 0$. This problem has a closed-form solution in which $u_1$ and $u_2$ are sinusoidal, and a representative solution is shown in~\cref{fig:example_sinusoid}. In this work, we use a fixed initial condition of $\mathbf{u}_0 = [0, \pi/2]^T$ and a frequency of $\omega=1$.

When being solved by a \gls{PINN}, the complexity of the \gls{ODE} is determined by the size of the time domain $T$. Although the solution is periodic, we will show that \glspl{PINN} fail to solve this \gls{ODE} as $T$ increases in~\cref{sec:errors}.

\subsection{Related work}\label{sec:related_work}

\glspl{PINN} were introduced in~\cite{raissi2019physics}, and there has been a great deal of follow-on research. We focus on work that considers the \gls{PINN} learning problem and recommend~\cite{karniadakis2021physics} as an overview of the broader space of physics-informed \gls{ML}.

Because the \gls{PINN} formulation is end-to-end differentiable, it is easily modified---e.g., for uncertainty estimation of components~\cite{yang2021b}. Although a given trained network $\hat{\mathbf{u}}_w$ is valid only for a single instantiation of the \gls{DE}, transfer learning approaches~\cite{arthurs2021activetraining} were shown to enable generalization over some spaces of parametrically-related equations.

\cite{wang2021understanding} showed that standard \glspl{PINN} can fail to solve simple \glspl{PDE} like the 2D Helmholtz equation with a known analytic solution. The \gls{PINN} struggled with fitting this problem's boundary condition, which the authors relate to underlying stiffness in the dynamics of the \gls{PINN} training process. They suggest several heuristic strategies for mitigating this stiffness, including a learning rate annealing process and a novel network architecture. Similarly,~\cite{Maddu2022dirichlet} use gradient uncertainties to adjust learning rates for \glspl{PINN}.

In follow-on work \cite{wang2022ntk} analyzed the dynamics of the \gls{PINN} learning problem through a \gls{NTK}~\cite{Jacot2018ntk} perspective and found that components of the \gls{PINN} loss function had different convergence rates. Most recently,~\cite{Wang2022causality} proposed a time-weighted scheme for training \glspl{PINN} that attempts to preserve physical causality in \glspl{DE}, and~\cite{Daw2022sampling} analyzed how \glspl{PINN} can fail to propgate solutions across large time-domains.


\cite{mclenny2020adaptive} observed that different regions of the domain can be harder to solve and introduced a set of attention weights to allow the network to adaptively focus on difficult regions of the domain. In a later addition to their paper, they built from~\cite{wang2022ntk} and conducted a \gls{NTK} analysis of this modified loss function.

\cite{krishnapriyan2021characterizing} focused instead on the difficulties that follow from the presence of differential operators in the \gls{PINN} loss function. This operator can lead to a poorly-conditioned loss function (as measured by local smoothness), which impairs the \gls{PINN}'s ability to model simple \glspl{PDE} like a reaction-diffusion model. \cite{krishnapriyan2021characterizing} also used Hessian-based information to analyze \glspl{PINN}--- specifically, they perturbed the \gls{PINN} loss in directions defined by the top two principal eigenvectors of the Hessian. \cite{wang2021understanding} tracked the principal eigenvalue of the Hessian to define a time scale of training. Outside of work with~\glspl{PINN},~\cite{yao2020pyhessian} analyze Hessian traces and spectral densities of feed-forward neural networks for computer vision problems, and~\cite{Yao2021adahessian} proposes a training scheme that uses an approximated Hessian diagonal to account for single-objective loss function curvature.

Similarly to our work,~\cite{Ji2021pinns_chemistry} noted that the \glspl{PINN} defined by \cite{raissi2019physics} struggle when attempting to model stiff systems of \glspl{ODE} for chemical kinetics models. They showed that the quasi-steady-state assumption, used to reduce the stiffness of a problem by assuming a zero concentration change rate for certain variables, can improve the accuracy of \glspl{PINN} when the reduced system is used as the \gls{PINN} loss $L(w)$. This approach showed promise for stiff systems, but will likely not mitigate the general challenge of \gls{ODE} system complexity with increased problem size and coupling.

Although the classic \gls{PINN} focuses on the strong formulation of a \gls{DE}, ~\cite{kharazmi2019variational} instead solved the weak formulation of a \gls{DE}. Other extensions to the \gls{PINN} learning problem came from~\cite{yang2020adversarialpinn}, who incorporated a generative adversarial component into their \gls{PINN}.

One of our benchmarks (\cref{sec:sinusoids}) shows that the accuracy of a \gls{PINN} breaks down as the size of the time domain $T$ increases. Prior work has considered similar settings for \glspl{PINN}~\cite{krishnapriyan2021characterizing,Meng2020parareal} and operator-learning approaches~\cite{Wang2021time_integration}, albeit focusing more on complicated \glspl{PDE}. We show here that the challenge of using \glspl{PINN} for long time-horizon forecasting exists even in the simplest \glspl{ODE}.

\section{Results}\label{sec:results}

\subsection{Evaluation procedure}\label{sec:procedure}

In~\cref{tab:hyperparameters} in~\cref{sec:supp-tables}, we show the network configurations and hyperparameters considered when training \glspl{PINN} for each benchmark. All hidden layers had the same number of units (64 or 128) depending on the particular architecture, and ResNet models incorporate skip connections~\cite{He2016residual} 
between hidden layers. These network sizes are typical of those used in \gls{PINN} literature (e.g.,~\cite{Wang2022causality,mclenny2020adaptive,Maddu2022dirichlet}, which primarily consider networks with 4-5 layers and 20-100 hidden units).

We selected training points across the time domain as $\mathcal{T} = \{T/D, 2T/D, \hdots, (D-1)T/D, T\}$, evaluation points as the midpoints of $\mathcal{T}$: $\mathcal{T}' = \{(3T)/(2D), \hdots, (2D-1)T / (2D)\}$, and for each benchmark/complexity value pair we trained a total of 48 \gls{PINN} configurations. All networks used $\tanh$ activation functions to ensure smoothness in the loss, and~\cref{tab:system_parameters} in~\cref{sec:supp-tables} includes benchmark-specific parameters.


\subsection{Evaluation metrics}\label{sec:metrics}
We evaluate \gls{PINN} solutions via several methods. The primary approach is with the relative $\ell^2$ error of the \gls{PINN} solution $\mathbf{\hat{u}}_w$ at the end of training vs. a solution $\mathbf{u}$ obtained with a classical \gls{ODE} solver:
$$\mathrm{RelError}_{\mathcal{T'}}(\mathbf{u}, \mathbf{\hat{u}}_w) = \sqrt{ \frac{\sum_{t' \in \mathcal{T}'} ||\mathbf{u}(t') - \mathbf{\hat{u}}_w(t')||^2}{\sum_{t' \in \mathcal{T}'} ||\mathbf{u}(t')||^2}},$$
where $\mathcal{T}' \subseteq (0, T)$ is a set of evaluation collocation points distinct from the training points $\mathcal{T}$.\footnote{Note that, in the forward prediction \gls{PINN} formulation, neither $\mathrm{RelError}_\mathcal{T}$ nor $\mathrm{RelError}_{\mathcal{T}'}$ are used as learning signals during training.} Classical solver solutions are obtained with the \texttt{solve\_ivp} function in \texttt{scipy.integrate} using \texttt{RK45}, the explicit Runge-Kutta method of order 5(4) (i.e., fourth-order accuracy with fifth-order local extrapolation)~\cite{Dormand1980rk45}.
In addition, we consider relative initial condition error
$$\mathrm{RelError}_0(\mathbf{u}, \mathbf{\hat{u}}_w) = \frac{||\mathbf{u}_0 - \hat{\mathbf{u}}_w(0)||}{||\mathbf{u}_0||}.$$

As discussed in~\cref{sec:pinns}, we track the complexity of the \gls{PINN} learning problem with the Laplacian of components of the \gls{PINN} loss function. We also demonstrate that our notion of benchmark ``complexity'' or ``difficulty'' correlates with this measure of complexity for some benchmarks. 

\subsection{Relative error analysis}\label{sec:errors}

\begin{figure*}
    \centering
    \includegraphics[width=0.8\linewidth]{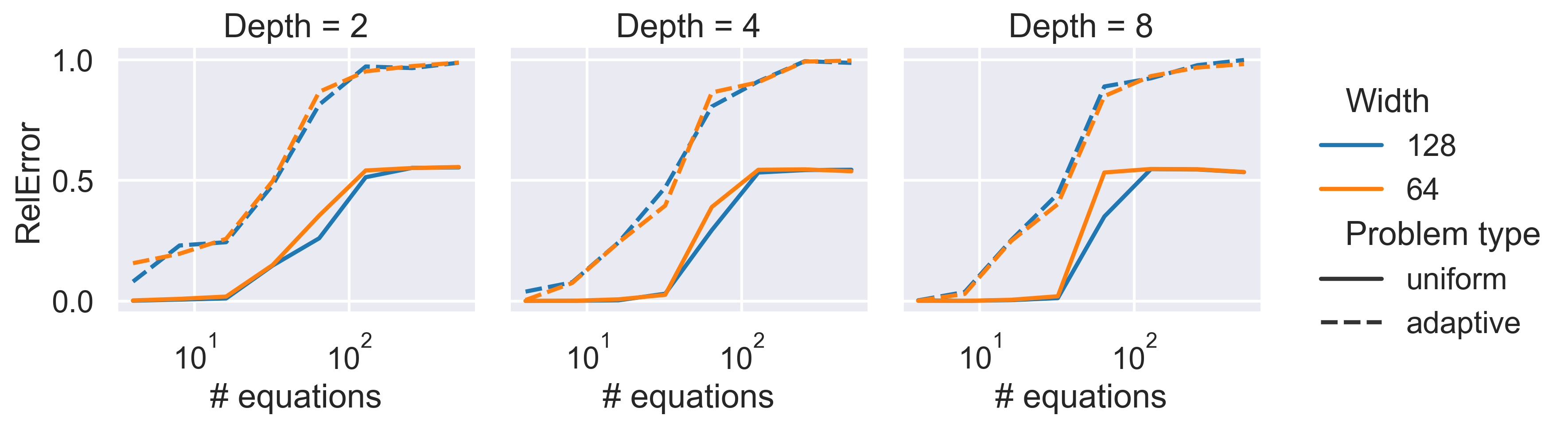}
    \caption{In the heat equation benchmark, all \glspl{PINN} fail to solve the problem as dimensionality $N$ increases. Adaptive weighting in the domain~\cite{mclenny2020adaptive} performs worse than uniform weighting---the weights $\lambda$ cannot mitigate the increase in ill-conditionedness of the problem. Interestingly, increasing network depth and width have little effect on the final relative error. Here we show relative error results for \glspl{MLP} initiated with a learning rate of $10^{-3}$; results for other \gls{PINN} configurations are comparable or worse.
    }
    \label{fig:heat_eqn_errors}
\end{figure*}

In~\cref{fig:heat_eqn_errors}, we show that \glspl{PINN} fail to solve the discretized heat equation problem (\cref{sec:heat}) as the discretization size $N$ increases. This happens across variations in network architecture and \gls{PINN} formulation. Despite our normalization of the residual loss by the norm of the discretized differential operator, the uniform \gls{PINN} formulation (\cref{eq:base_pinn}) fails to solve the problem, and the adaptive \gls{PINN} (\cref{eq:weighted_pinn}) performs worse.


\begin{figure*}
    \centering
    \includegraphics[width=0.8\linewidth]{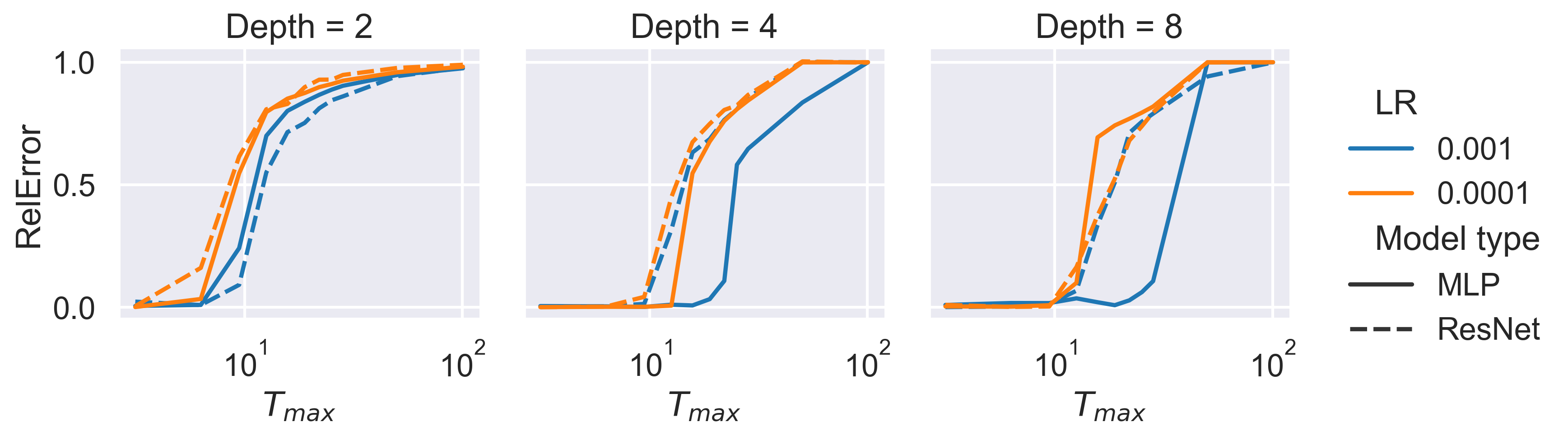}
    \caption{For the \gls{SHM} benchmark and network depths of 2, 4, and 8, all \glspl{PINN} increasingly fail to solve the problem as the maximum time $T$ increases. Depth reduces relative error, although error is still dependent on other hyperparameters---the high learning rate of 0.001 and a MLP perform the best. Here we show error results for models under the adaptive weighting \gls{PINN} formulation with a width of 64 units; results for other \gls{PINN} configurations are comparable or worse. Upper bounds on $T$ are of form $T_{max}\pi$ when actually training the \glspl{PINN}. 
    Precise values of maximum time are given in~\cref{tab:system_parameters} in~\cref{sec:supp-tables}.
    }
    \label{fig:shm_errors}
\end{figure*}

In~\cref{fig:shm_errors}, we show that \glspl{PINN} fail to solve the \gls{SHM} problem (\cref{sec:sinusoids}) as the maximum time $T$ of the domain increases. This happens despite our variations in network architecture and \gls{PINN} formulation, as well as the fact that we increase the number of training points $D$ as the maximum time $T$ increases (\cref{sec:supp-tables}, \cref{tab:system_parameters}). All configurations are able to effectively learn the problem's initial condition---in~\cref{fig:shm_ic_error} in~\cref{sec:supp-figs}, we show that the average relative errors for the initial condition do not exceed 0.02. In this problem, the adaptive weighting scheme~\cite{mclenny2020adaptive} attains the lowest relative error but is still dependent on other hyperparameters.

In~\cref{fig:shm_errors} (\gls{SHM}) but not~\cref{fig:heat_eqn_errors} (heat equation), deeper networks were able to solve more complex problems---although the high-end of time domain size still results in a failed \gls{PINN}. This is likely not an indication that increased network depth is a solution, as the underlying phenomena has a simple periodic structure with fixed complexity. Thus, the \gls{PINN} is not discovering the periodic structure in \gls{SHM} and can only learn the \gls{DE}'s behavior with very high-capacity networks. This works in this \gls{SHM} setting, but will not scale to complex long-time horizon phenomena.

\subsection{Laplacian analysis}\label{sec:trace_analysis}

\begin{figure*}
    \centering
    \includegraphics[width=0.6\linewidth]{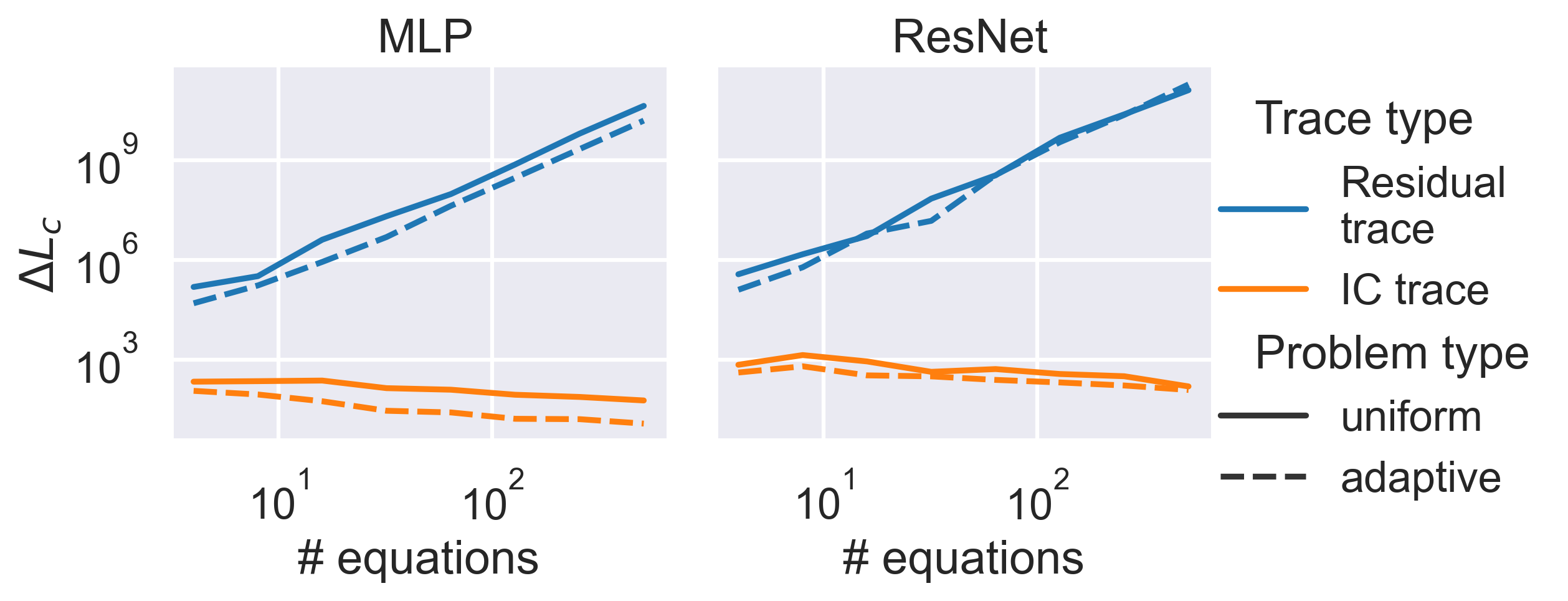}
    \caption{Analysis of the Laplacian $\Delta L_c$ of the components of the heat equation loss function, where the residual trace has been normalized by the condition number of the discretized differential operator, and the initial condition trace has been normalized by the number of equations. Under this normalization, the Laplacian of the initial condition loss slightly decreases as the number of equations increases, but the Laplacian of the residual loss increases rapidly. Laplacians of the uniform and adaptive \gls{PINN} formulations remain comparable, despite the latter performing much worse with respect to relative error (\cref{fig:heat_eqn_errors}). Results are shown for models with 8 hidden layers of 128 units, trained with an initial learning rate of $10^{-3}$.
    }
    \label{fig:heat_equation_trace_analysis}
\end{figure*}

In~\cref{fig:heat_equation_trace_analysis}, we analyze the Laplacian $\Delta L_c$ of the components of the heat equation loss function. We seek trends in the Laplacian as the number of equations varies; thus, we normalize the Laplacian of the residual by the condition number of the discretized differential operator and the Laplacian of the initial condition error by the number of equations. Even with this normalization, there remains a positive correlation between the number and equations and the Laplacian of the residual---the \gls{PINN} formulation cannot handle the ill-conditioning of the heat equation benchmark.

\begin{figure*}
    \centering
    \includegraphics[width=0.8\linewidth]{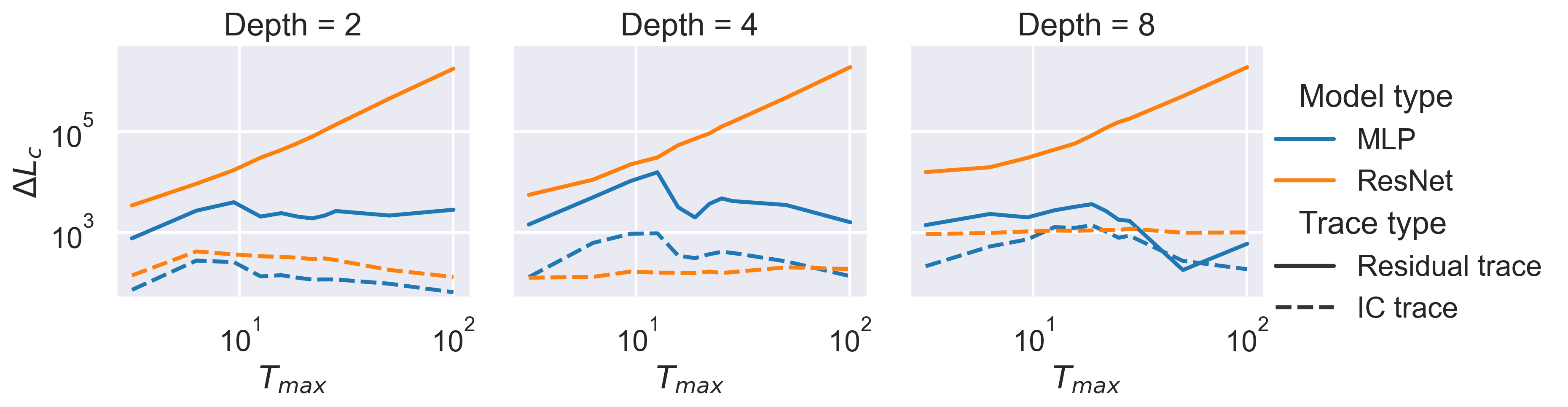}
    \caption{Analysis of the Laplacian $\Delta L_c$ of components of the \gls{SHM} loss function. The Laplacian of the initial condition loss remains fairly constant as network configurations and maximum time $T$ change, reflecting how the initial condition error for this problem is easily satisfied. Interestingly, the Laplacian of \glspl{PINN} with ResNet networks is consistently much larger than those with \glspl{MLP}, which aligns with the findings of~\cite{yao2020pyhessian} but contradicts findings of~\cite{Li2018curvature}; both of whom consider computer vision problems. We show results for the uniform \gls{PINN} loss formulation, initiated with a learning rate of $10^{-3}$; other results are comparable.}
    \label{fig:shm_trace_analysis}
\end{figure*}

In~\cref{fig:shm_trace_analysis}, we analyze the Laplacian $\Delta L_c$ of the components of the \gls{SHM} loss function. For ResNet architectures, the Laplacian of the residual loss does increase as the maximum time $T$ also increases; however, this increase is not also found for \gls{MLP} architectures. 

\section{Discussion}
In this paper, we demonstrate that, in a pair of benchmarks, the relative error of a \gls{PINN} falters as problem complexity increases. Our results serve as a cautionary tale in that the ability of modern \gls{ML} to ``reason'' in the language of physical laws still remains challenging. 

Our experiments can be expanded on in several ways. First, we only consider forward problems, in the assumption that inverse problems would require robust forward prediction capability. This seems necessary, but may not be sufficient for successful solutions to inverse problems. Second, our current results may benefit from a larger sweep of model hyperparameters---which could provide additional insight into \gls{PINN} behavior and increase confidence in our results. Finally, our choice of two benchmarks is focused on \gls{PINN} performance for \glspl{ODE}---as we discuss below, a broader benchmark suite is likely needed.

The two benchmarks chosen for this paper are simple and representative of actual \gls{DE} modeling problems encountered in scientific applications. We identify Hessian-based metrics as being sometimes representative of a \gls{PINN}'s ability to solve a problem, but further work remains in a devising a more general classification of \glspl{PINN} and \glspl{DE}. This can assist in evaluating novel approaches for \gls{PINN} architectures, loss functions, and training schemes. Our benchmarks can easily be made more extensive. For example, more than one dimension of complexity can be varied, and a variety of initial conditions can be specified. In addition, similar benchmarks can be developed for \glspl{PDE}. We believe that a robust suite of benchmarks for physics-informed \gls{ML} models is needed to facilitate reproducibility and to aid validation of models for difficult problems---such as large, stiff, and multi-scale systems, as well as inverse problems.



\section*{Acknowledgements}

This work was supported by internal research and development funding from the Homeland Protection Mission Area of the Johns Hopkins University Applied Physics Laboratory. Thanks to Christine D. Piatko and I-Jeng Wang for assistance in editing and improving this work.


\bibliography{references}

\begin{thebibliography}{36}
\providecommand{\natexlab}[1]{#1}

\bibitem[{Arthurs and King(2021)}]{arthurs2021activetraining}
Arthurs, C.~J.; and King, A.~P. 2021.
\newblock Active {T}raining of {P}hysics-{I}nformed {N}eural {N}etworks to
  {A}ggregate and {I}nterpolate {P}arametric {S}olutions to the
  {N}avier-{S}tokes {E}quations.
\newblock \emph{Journal of Computational Physics}, 438: 110364.

\bibitem[{Baydin et~al.(2017)Baydin, Pearlmutter, Radul, and
  Siskind}]{Baydin2017autodiff}
Baydin, A.~G.; Pearlmutter, B.~A.; Radul, A.~A.; and Siskind, J.~M. 2017.
\newblock {Automatic Differentiation in Machine Learning: A Survey}.
\newblock \emph{J. Mach. Learn. Res.}, 18(1): 5595–5637.

\bibitem[{Brandstetter, Worrall, and Welling(2022)}]{brandstetter2022message}
Brandstetter, J.; Worrall, D.~E.; and Welling, M. 2022.
\newblock {Message Passing Neural {PDE} Solvers}.
\newblock In \emph{International Conference on Learning Representations}.

\bibitem[{Chow, Mallet-Paret, and Van~Vleck(1996)}]{chow1996dynamics}
Chow, S.-N.; Mallet-Paret, J.; and Van~Vleck, E.~S. 1996.
\newblock {Dynamics of Lattice Differential Equations}.
\newblock \emph{International Journal of Bifurcation and Chaos}, 06(09):
  1605--1621.

\bibitem[{Daw et~al.(2022)Daw, Bu, Wang, Perdikaris, and
  Karpatne}]{Daw2022sampling}
Daw, A.; Bu, J.; Wang, S.; Perdikaris, P.; and Karpatne, A. 2022.
\newblock Mitigating Propagation Failures in PINNs using Evolutionary Sampling.
\newblock Doi:10.48550/ARXIV.2207.02338.

\bibitem[{Dormand and Prince(1980)}]{Dormand1980rk45}
Dormand, J.; and Prince, P. 1980.
\newblock {A Family of Embedded {R}unge-{K}utta Formulae}.
\newblock \emph{Journal of Computational and Applied Mathematics}, 6(1):
  19--26.

\bibitem[{He et~al.(2016)He, Zhang, Ren, and Sun}]{He2016residual}
He, K.; Zhang, X.; Ren, S.; and Sun, J. 2016.
\newblock {Deep Residual Learning for Image Recognition}.
\newblock In \emph{2016 IEEE Conference on Computer Vision and Pattern
  Recognition (CVPR)}, 770--778.

\bibitem[{Jacot, Gabriel, and Hongler(2018)}]{Jacot2018ntk}
Jacot, A.; Gabriel, F.; and Hongler, C. 2018.
\newblock {Neural Tangent Kernel: Convergence and Generalization in Neural
  Networks}.
\newblock In \emph{{Advances in Neural Information Processing Systems}},
  volume~31.

\bibitem[{Ji et~al.(2021)Ji, Qiu, Shi, Pan, and Deng}]{Ji2021pinns_chemistry}
Ji, W.; Qiu, W.; Shi, Z.; Pan, S.; and Deng, S. 2021.
\newblock {Stiff-{PINN}: Physics-Informed Neural Network for Stiff Chemical
  Kinetics}.
\newblock \emph{The Journal of Physical Chemistry A}, 125(36): 8098--8106.

\bibitem[{Karniadakis et~al.(2021)Karniadakis, Kevrekidis, Lu, Perdikaris,
  Wang, and Yang}]{karniadakis2021physics}
Karniadakis, G.~E.; Kevrekidis, I.~G.; Lu, L.; Perdikaris, P.; Wang, S.; and
  Yang, L. 2021.
\newblock {Physics-Informed Machine Learning}.
\newblock \emph{Nature Reviews Physics}, 3(6): 422--440.

\bibitem[{Kharazmi, Zhang, and Karniadakis(2021)}]{kharazmi2019variational}
Kharazmi, E.; Zhang, Z.; and Karniadakis, G.~E. 2021.
\newblock {hp-VPINNs: Variational Physics-Informed Neural Networks with Domain
  Decomposition}.
\newblock \emph{Computer Methods in Applied Mechanics and Engineering}, 374:
  113547.

\bibitem[{Kingma and Ba(2014)}]{Kingma2014adam}
Kingma, D.~P.; and Ba, J. 2014.
\newblock {Adam: A Method for Stochastic Optimization}.
\newblock Doi:10.48550/ARXIV.1412.6980.

\bibitem[{Krishnapriyan et~al.(2021)Krishnapriyan, Gholami, Zhe, Kirby, and
  Mahoney}]{krishnapriyan2021characterizing}
Krishnapriyan, A.; Gholami, A.; Zhe, S.; Kirby, R.; and Mahoney, M.~W. 2021.
\newblock {Characterizing Possible Failure Modes in Physics-Informed Neural
  Networks}.
\newblock \emph{Advances in Neural Information Processing Systems}, 34.

\bibitem[{Li et~al.(2018)Li, Farkhoor, Liu, and Yosinski}]{Li2018curvature}
Li, C.; Farkhoor, H.; Liu, R.; and Yosinski, J. 2018.
\newblock {Measuring the Intrinsic Dimension of Objective Landscapes}.
\newblock In \emph{International Conference on Learning Representations}.

\bibitem[{Lu et~al.(2021)Lu, Meng, Mao, and Karniadakis}]{lu2021deepxde}
Lu, L.; Meng, X.; Mao, Z.; and Karniadakis, G.~E. 2021.
\newblock {{DeepXDE}: {A} Deep Learning Library for Solving Differential
  Equations}.
\newblock \emph{SIAM Review}, 63(1): 208--228.

\bibitem[{Maddu et~al.(2022)Maddu, Sturm, Müller, and
  Sbalzarini}]{Maddu2022dirichlet}
Maddu, S.; Sturm, D.; Müller, C.~L.; and Sbalzarini, I.~F. 2022.
\newblock {Inverse Dirichlet Weighting Enables Reliable Training of Physics
  Informed Neural Networks}.
\newblock \emph{Machine Learning: Science and Technology}, 3(1): 015026.

\bibitem[{McClenny and Braga-Neto(2020)}]{mclenny2020adaptive}
McClenny, L.; and Braga-Neto, U. 2020.
\newblock {Self-Adaptive Physics-Informed Neural Networks using a Soft
  Attention Mechanism}.
\newblock Doi:10.48550/ARXIV.2009.04544.

\bibitem[{McClenny, Haile, and Braga-Neto(2021)}]{mcclenney2021tensordiffeq}
McClenny, L.~D.; Haile, M.~A.; and Braga-Neto, U.~M. 2021.
\newblock {TensorDiffEq: Scalable Multi-{GPU} Forward and Inverse Solvers for
  Physics Informed Neural Networks}.
\newblock Doi:10.48550/ARXIV.2103.16034.

\bibitem[{Meng et~al.(2020)Meng, Li, Zhang, and Karniadakis}]{Meng2020parareal}
Meng, X.; Li, Z.; Zhang, D.; and Karniadakis, G.~E. 2020.
\newblock {{PPINN}: Parareal Physics-Informed Neural Network for Time-Dependent
  PDEs}.
\newblock \emph{Computer Methods in Applied Mechanics and Engineering}, 370:
  113250.

\bibitem[{Noschese, Pasquini, and Reichel(2013)}]{Noschese2013toeplitz}
Noschese, S.; Pasquini, L.; and Reichel, L. 2013.
\newblock {Tridiagonal {T}oeplitz Matrices: Properties and Novel Applications}.
\newblock \emph{Numerical Linear Algebra with Applications}, 20(2): 302--326.

\bibitem[{Pearlmutter(1994)}]{pearlmutter1994hessian}
Pearlmutter, B.~A. 1994.
\newblock {{Fast Exact Multiplication by the Hessian}}.
\newblock \emph{Neural Computation}, 6(1): 147--160.

\bibitem[{Raissi, Perdikaris, and Karniadakis(2019)}]{raissi2019physics}
Raissi, M.; Perdikaris, P.; and Karniadakis, G.~E. 2019.
\newblock {Physics-Informed Neural Networks: {A} Deep Learning Framework for
  Solving Forward and Inverse Problems Involving Nonlinear Partial Differential
  Equations}.
\newblock \emph{Journal of Computational Physics}, 378: 686--707.

\bibitem[{Shin(2020)}]{shin2020convergence}
Shin, Y. 2020.
\newblock {On the Convergence of Physics Informed Neural Networks for Linear
  Second-Order Elliptic and Parabolic Type {PDEs}}.
\newblock \emph{Communications in Computational Physics}, 28(5): 2042--2074.

\bibitem[{Singhal et~al.(2021)Singhal, Tuza, Sun, and
  Murray}]{singhal2021txtlsim}
Singhal, V.; Tuza, Z.~A.; Sun, Z.~Z.; and Murray, R.~M. 2021.
\newblock {{A MATLAB Toolbox for Modeling Genetic Circuits in Cell-Free
  Systems}}.
\newblock \emph{Synthetic Biology}, 6(1).

\bibitem[{Verwer and Sanz-Serna(1984)}]{Verwer1984methodoflines}
Verwer, J.; and Sanz-Serna, J. 1984.
\newblock {Convergence of Method of Lines Approximations to Partial
  Differential Equations}.
\newblock \emph{Computing}, 33(3-4): 297--313.

\bibitem[{Wang and Perdikaris(2021)}]{Wang2021time_integration}
Wang, S.; and Perdikaris, P. 2021.
\newblock {Long-time Integration of Parametric Evolution Equations with
  Physics-Informed DeepONets}.
\newblock Doi: 10.48550/ARXIV.2106.05384.

\bibitem[{Wang, Sankaran, and Perdikaris(2022)}]{Wang2022causality}
Wang, S.; Sankaran, S.; and Perdikaris, P. 2022.
\newblock {Respecting Causality is all you Need for Training Physics-Informed
  Neural Networks}.

\bibitem[{Wang, Teng, and Perdikaris(2021)}]{wang2021understanding}
Wang, S.; Teng, Y.; and Perdikaris, P. 2021.
\newblock {Understanding and Mitigating Gradient Flow Pathologies in
  Physics-Informed Neural Networks}.
\newblock \emph{SIAM Journal on Scientific Computing}, 43(5): A3055--A3081.

\bibitem[{Wang, Yu, and Perdikaris(2022)}]{wang2022ntk}
Wang, S.; Yu, X.; and Perdikaris, P. 2022.
\newblock {When and Why {PINN}s Fail to Train: A Neural Tangent Kernel
  Perspective}.
\newblock \emph{Journal of Computational Physics}, 449: 110768.

\bibitem[{Willard et~al.(2020)Willard, Jia, Xu, Steinbach, and
  Kumar}]{willard2020integrating}
Willard, J.; Jia, X.; Xu, S.; Steinbach, M.; and Kumar, V. 2020.
\newblock {Integrating Scientific Knowledge with Machine Learning for
  Engineering and Environmental Systems}.

\bibitem[{Yang, Meng, and Karniadakis(2021)}]{yang2021b}
Yang, L.; Meng, X.; and Karniadakis, G.~E. 2021.
\newblock {{B-PINNs}: {B}ayesian Physics-Informed Neural Networks for Forward
  and Inverse {PDE} Problems with Noisy Data}.
\newblock \emph{Journal of Computational Physics}, 425: 109913.

\bibitem[{Yang, Zhang, and Karniadakis(2020)}]{yang2020adversarialpinn}
Yang, L.; Zhang, D.; and Karniadakis, G.~E. 2020.
\newblock {Physics-Informed Generative Adversarial Networks for Stochastic
  Differential Equations}.
\newblock \emph{SIAM Journal on Scientific Computing}, 42(1): A292--A317.

\bibitem[{Yao et~al.(2020)Yao, Gholami, Keutzer, and
  Mahoney}]{yao2020pyhessian}
Yao, Z.; Gholami, A.; Keutzer, K.; and Mahoney, M.~W. 2020.
\newblock {Pyhessian: {N}eural Networks Through the Lens of the {H}essian}.
\newblock In \emph{2020 IEEE International Conference on Big Data (Big Data)},
  581--590. IEEE.

\bibitem[{Yao et~al.(2021)Yao, Gholami, Shen, Mustafa, Keutzer, and
  Mahoney}]{Yao2021adahessian}
Yao, Z.; Gholami, A.; Shen, S.; Mustafa, M.; Keutzer, K.; and Mahoney, M. 2021.
\newblock {ADAHESSIAN: An Adaptive Second Order Optimizer for Machine
  Learning}.
\newblock \emph{Proceedings of the AAAI Conference on Artificial Intelligence},
  35(12): 10665--10673.

\bibitem[{Zogheib et~al.(2021)Zogheib, Tohidi, Baskonus, and
  Cattani}]{Zogheib2021burgers}
Zogheib, B.; Tohidi, E.; Baskonus, H.~M.; and Cattani, C. 2021.
\newblock {Method of Lines for Multi-Dimensional Coupled Viscous {Burgers'}
  Equations via Nodal {Jacobi} Spectral Collocation Method}.
\newblock \emph{Physica Scripta}, 96(12): 124011.

\bibitem[{Zubov et~al.(2021)Zubov, McCarthy, Ma, Calisto, Pagliarino, Azeglio,
  Bottero, Luján, Sulzer, Bharambe, Vinchhi, Balakrishnan, Upadhyay, and
  Rackauckas}]{zubov2021neuralpde}
Zubov, K.; McCarthy, Z.; Ma, Y.; Calisto, F.; Pagliarino, V.; Azeglio, S.;
  Bottero, L.; Luján, E.; Sulzer, V.; Bharambe, A.; Vinchhi, N.; Balakrishnan,
  K.; Upadhyay, D.; and Rackauckas, C. 2021.
\newblock {{NeuralPDE}: Automating Physics-Informed Neural Networks ({PINNs})
  with Error Approximations}.
\newblock Doi: 10.48550/ARXIV.2107.09443.

\end{thebibliography}

\newpage
\appendix
\onecolumn

\section{Details on the heat equation}\label{sec:heat_deets}

The continuous-space formulation of the heat equation is:
\begin{eqnarray*}
    \partial_t u &=& \partial_{xx} u\\
    u(x, 0) &=& g(x)\\
    u(0, t) &=& u_L(t)\\
    u(1, t) &=& u_R(t)\\
    (x, t) &\in& (0, 1) \times (0, T),
\end{eqnarray*}

where $g$ specifies the initial condition, and $u_L$ and $u_R$ specify the boundary conditions on the left and right side, respectively, of the spatial domain $(0, 1)$. We use an initial condition of $g(x) = \sin(2\pi x)+1$ and boundary conditions of $u_L(t) = u_R(t) = 1$.

The central finite difference approximation to $\partial_{xx}$ is:
$$\partial_{xx} u(x_n, t) = \frac{u(x_{n+1}, t) - 2u(x_n, t) + u(x_{n-1}, t)}{(\Delta x)^2},$$
which yields the system of linear equations described in~\cref{sec:heat}.

\section{Supplemental tables}\label{sec:supp-tables}
\begin{table}[h]
    \centering
    \begin{tabular}{c|c}
    Hyperparameter          &   Value(s) used\\\hline
    \# Layers               &   2, 4, 8\\   
    \# Hidden units         &   64, 128\\
    Initial learning rate   &   $10^{-3}, 10^{-4}$\\
    Network type            &   MLP, ResNet\\
    \gls{PINN} formulation  &   Uniform, Adaptive\\
    Activation function     &   $\tanh$\\
    Masking function $\mu$  &   Sigmoid\\
    \# Adam iterations      &   10,241\\
    \# Training points $D$ (SHM) &$256T/\pi$\\
    \# Training points $D$ (heat equation) &   1,024
    \end{tabular}
    \caption{Hyperparameters searched over when solving \gls{PINN} problems. For the \gls{PINN} formulation, Uniform is the standard \gls{PINN} formulation~\cref{eq:base_pinn}, and Adaptive is the adaptive-weighted \gls{PINN} formulation~\cref{eq:weighted_pinn}. The heat equation benchmark uses a fixed number of training points, since their domains do not vary in size; the \gls{SHM} benchmark scales the number of training points with the time horizon $T$. 
    }
    \label{tab:hyperparameters}
\end{table}

\begin{table*}[h]
    \centering
    \begin{tabular}{c|c|c|c}
    Problem                 &   Complexity parameter                &   Values evaluated                    &   Other parameters \\\hline
    Simple harmonic motion  &   Maximum time                        &   $T=2^\ell \pi, \ell=0,1,\hdots,5$             &   $N=2$, $\nu_{\mathcal{I}} = 1$\\ 
    Heat equation           &   \# discretized points               &   $N$ = 4, 8, ..., 256, 512     &   $\nu_{\mathcal{I}} = ||\mathbf{A}||_2,T=0.1$\\
    \end{tabular}
    \caption{\gls{ODE} system parameters evaluated. For~\cref{sec:heat}, since the norm of $\mathbf{A}$ scales with the number of discretizations $N$ and the initial condition remains constant, we scale the initial condition error by $||\mathbf{A}||_2$ during training. For~\cref{sec:heat}, we choose the maximum time $T$ so that the solution has decayed to a constant value.}
    \label{tab:system_parameters}
\end{table*}

\newpage

\section{Supplemental Figures}\label{sec:supp-figs}


\begin{figure}[h]
    \centering
    \includegraphics[width=0.35\linewidth]{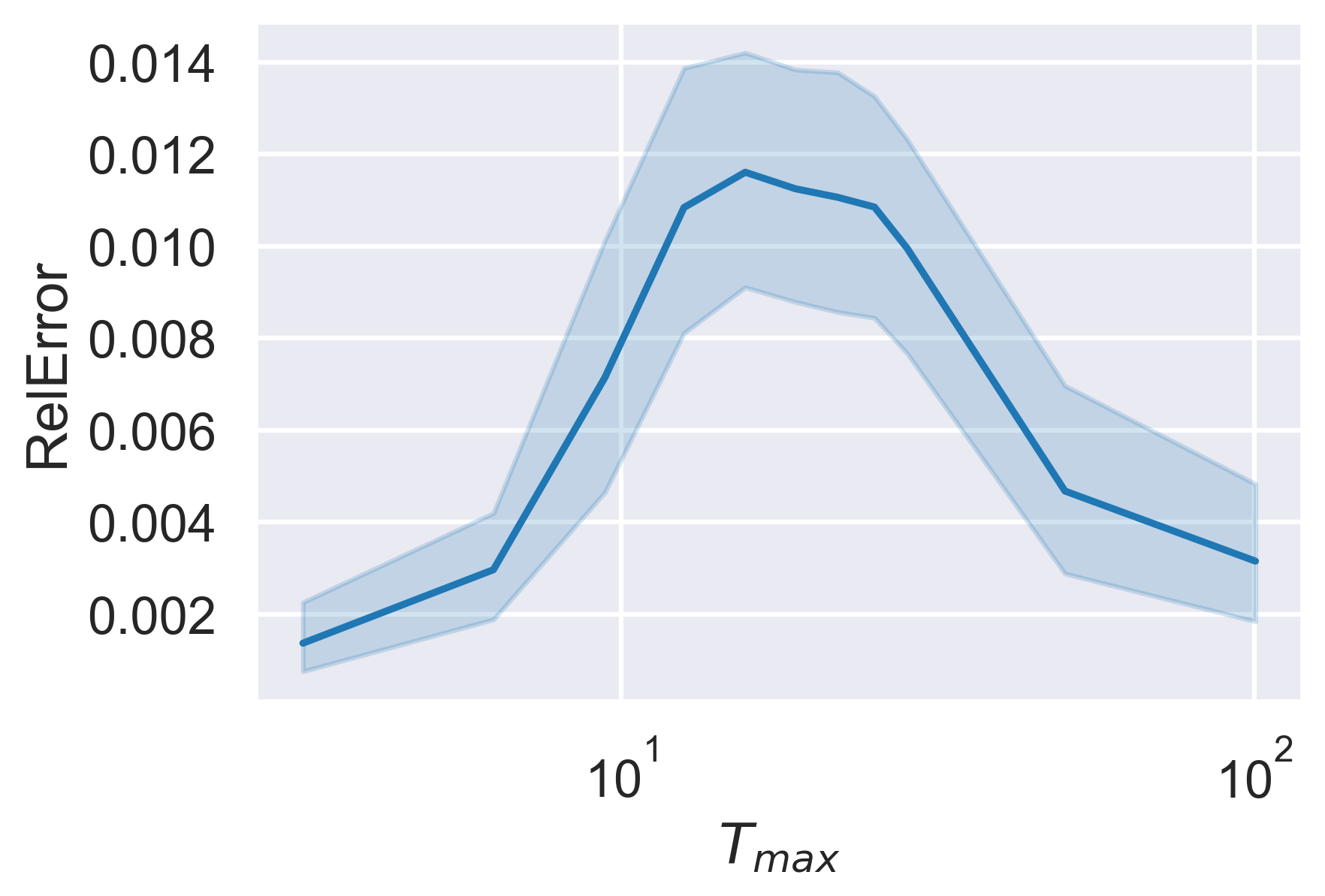}
    \caption{For the \gls{SHM} problem, all \glspl{PINN} are able to consistently minimize the relative initial condition error. Here we plot the final relative errors, averaged over all network configurations. Error bars indicate 95\% confidence intervals for the estimate of the mean. 
    }
    \label{fig:shm_ic_error}
\end{figure}

\begin{figure*}[h]
    \centering
    \includegraphics[width=0.9\linewidth]{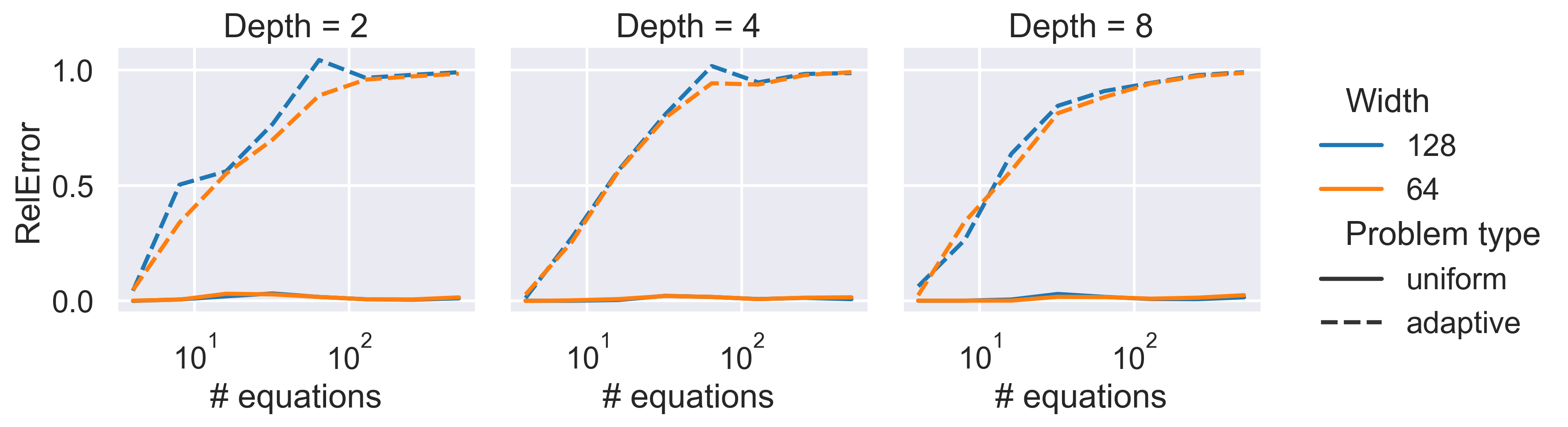}
    \caption{Even though \glspl{PINN} fail to solve the high-dimensional heat equation over the entire domain, the standard \gls{PINN} formulation is able to learn the initial condition.
    }
    \label{fig:heat_eqn_ic_errors}
\end{figure*}






\end{document}